# Graph-of-Thought: Utilizing Large Language Models to Solve Complex and Dynamic Business Problems

Ye Li
Asia-Pacific Research and Development Group
Microsoft
Beijing, China
jull@microsoft.com

*Abstract*—This paper presents Graph-of-Thought (GoT), a new model for workflow automation that enhances the flexibility and efficiency of Large Language Models (LLMs) in complex task execution. GoT advances beyond traditional linear and tree-like cognitive models with a graph structure that enables dynamic path selection. The open-source engine GoTFlow demonstrates the practical application of GoT, facilitating automated, data-driven decision-making across various domains. Despite challenges in complexity and transparency, GoTFlow's potential for improving business processes is significant, promising advancements in both efficiency and decision quality with continuous development.

*Keywords—Graph-of-Thought (GoT), Workflow Automation, Large Language Models (LLMs), Task Execution, Data-Driven Decision Making, Complexity Management*

## I. Introduction

Currently, with the rapid development of artificial intelligence and machine learning technologies, large language models (LLMs) have become a hot topic in the field of natural language processing. LLMs have demonstrated outstanding performance in various tasks such as text generation, translation, summarization, and question-answering systems.

However, LLMs still have limitations in handling complex, multi-step task processes. Traditional LLMs often encounter difficulties in generating coherent and logically consistent long texts, especially when they need to integrate information from multiple sources and perform deep reasoning.

The limitations of existing models like Chain-of-Thought (CoT) and Tree-of-Thought (ToT) stem from their linear and hierarchical structures, which struggle with complex, multi-dimensional problems. CoT is limited by its sequential reasoning, unable to explore multiple paths simultaneously, while ToT, despite allowing for branching, cannot model interconnected thought processes effectively.

To overcome these limitations, this paper proposes a new thinking model - Graph-of-Thought (GoT). Different from the CoT and ToT models, the proposed GoT model addresses these constraints by utilizing a graph structure for more dynamic, interconnected reasoning.

This approach surpasses CoT and ToT by enabling multi-threaded thought processes and integrating information from diverse sources, enhancing problem-solving accuracy and efficiency. GoT represents a significant leap forward, offering a more flexible, effective framework for complex task execution and workflow automation.

The main purpose of this paper is to introduce the concept and usage of Graph-of-Thought and demonstrate its value in practical applications through a workflow automation framework - GoTFlow.

GoTFlow is an open-source project that parses and executes large model workflows based on mind maps, automating the combination of a series of complex tasks. This paper will detail its architecture, working principles, and how to use it to achieve the transition from theory to practice.

Through this research, we aim to provide a more flexible and effective framework for solving complex problems and offer an innovative solution for automating workflows.

## II. Background

First, we need to clarify 2 concepts, which are Chain-of-Thought (CoT), and Tree-of-Thought (ToT).

Chain-of-Thought (CoT) and Tree-of-Thought (ToT) are both thinking models to construct prompts.

CoT presents the problem-solving process in a linear and step-by-step manner, and reveals how to proceed from the problem to the answer in a logical sequence, step by step. It is particularly suitable for tasks that require coherent reasoning, such as solving mathematical problems or complex inferences.

However, CoT is often unable to effectively integrate information in multi-branch tasks. A more complex model - Tree-of-Thought (ToT) was proposed.

ToT visualizes decision-making in a branching tree structure. Each branch represents a different reasoning path or solution, allowing for exploration of multiple outcomes. Although it captures a range of possibilities and their potential consequences, ToT still appears too rigid when facing intersecting and overlapping thinking paths.

To overcome the aforementioned limitations, we proposes a new model.

## III. GRAPH-OF-THOUGHT (GoT) MODEL

Graph-of-Thought (GoT) is an innovative thinking model that uses graph structures to define and execute tasks. Each "node" of thought is interconnected in multiple dimensions, providing not only linear and tree-like thinking paths but also creating more complex network structures.

This structure allows for the simultaneous existence of multiple thinking paths and decision points, thus more closely resembling the natural thinking process of humans.

Compared to CoT and ToT, the significant advantage of GoT lies in its flexibility and practicality.

By utilizing graph structures, GoT can more accurately simulate and execute multi-step, multi-condition workflows.

In the GoT model, each node can be a decision point, allowing for dynamic determination of subsequent paths based on the output of previous nodes.

This approach greatly enhances the ability to handle complex and variable tasks, making GoT particularly suitable for scenarios that require highly personalized and dynamic decision-making, such as complex data analysis, strategy formulation, project management, and other fields.

There are several ongoing efforts in the development of Graph-of-Thought frameworks, and this paper focuses on the implementation and application of Microsoft's Large Language Model in the proposed GoTFlow.

## IV. GOTFLOW: OPEN-SOURCE GRAPH-OF-THOUGHT WORKFLOW ENGINE

GoTFlow (https://github.com/microsoft/gotflow) is an open-source Graph-of-Thought workflow automation framework built on the Graph-of-Thought theory. It allows users to graphically define and execute task flows based on large language models (LLMs) for various business scenarios, thus achieving automated processing of complex task combinations.

Microsoft's implementation of the Graph-of-Thought model in GoTFlow leverages its powerful Large Language Model, which enables more accurate and efficient task execution and dynamic workflow adjustments.

The core module of GoTFlow is the Graph-of-Thought Flow Engine. Users define workflows through JSON format configuration files, which are then parsed and executed by the flow engine.

In GoTFlow, workflows are composed of two types of nodes which were linked with directed edges. The two types of nodes being Executor and Decision Maker.

*Executor*: Responsible for executing specific tasks. Each executor represents an operation or a set of operations, such as text segmentation, text integration, and prompting LLM tasks.

*Decision Maker*: Performing tasks and then determining the next direction of the workflow based on the output of the task execution. It can choose different execution paths based on complex logic or conditional expressions, similar to a branch (conditional) node in a control flow. This allows GoTFlow to not only execute fixed task sequences but also dynamically adapt to changing situations and requirements.

Essential functions of GoTFlow support the individual processing of nodes by executing tasks and determining the next steps in the workflow. They manage entire workflows, from handling single configurations to comprehensive workflow processing, emphasizing GoTFlow's capacity to manage complex workflows efficiently.

GoTFlow employs caching to store node outputs, facilitating their use in subsequent tasks. This supports complex, interdependent workflows. Robust error handling ensures workflow continuity, even in the face of errors, underscoring GoTFlow's reliability for managing large and complex workflows.

Through its unique architecture (refer to the blow diagram) and core components, GoTFlow not only provides powerful tools for executing complex tasks but also opens up new possibilities for automating and optimizing workflows. GoTFlow offers a novel solution for automating and optimizing workflows, providing unprecedented flexibility and efficiency in addressing business challenges across various applications.

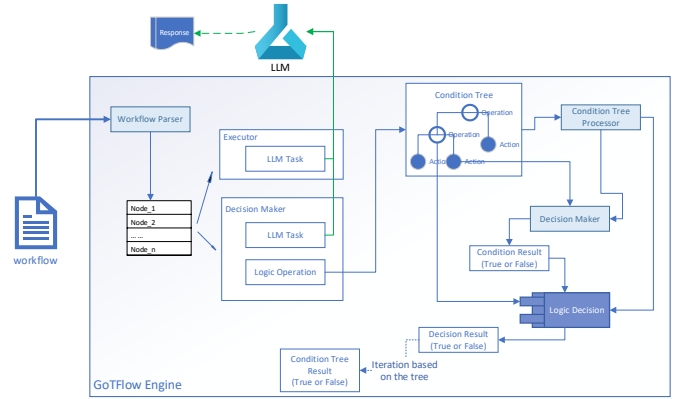

Fig. 1 Architecture of GoTFlow

## V. WORKFLOW DESIGN OF GOTFLOW

GoTFlow introduces a domain-specific language (DSL) to design and define workflows. GoTFlow DSL is based on JSON format because JSON has a clear structure and is easy to understand and write. In GoTFlow DSL, users can define various nodes in the workflow, directed relationships between nodes, and input/output, data processing, conditional judgments, or interactions with external services for each node.

Once the workflow is defined and configured, GoTFlow's workflow engine can automatically parse and execute these tasks, achieving the automation of the entire workflow.

To better understand GoTFlow's workflow design and execution process, here is a simple example:

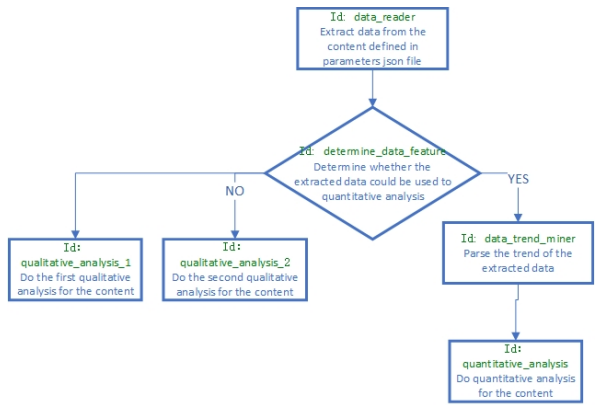

Fig. 2. Example of a GoTFlow's workflow.

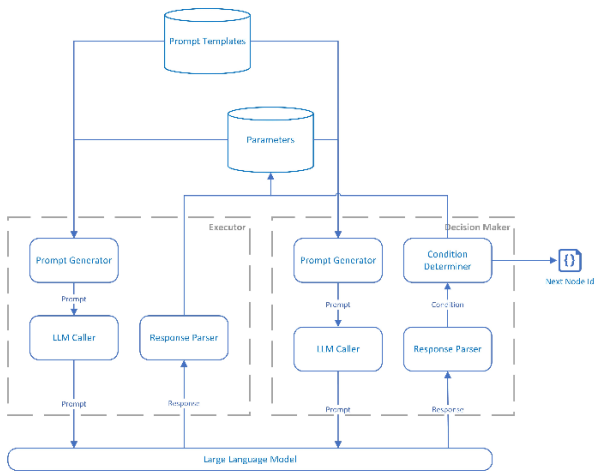

Fig. 3. Detailed execution logic of Executor and Decision Maker.

Defining Workflow Structure: Suppose we need to build a workflow for automatically collecting data, determining whether the collected data can be subjected to quantitative analysis, and if feasible, performing data mining and quantitative analysis; otherwise, performing qualitative analysis.

In this workflow, we need to define six nodes: an executor for data collection; a decision maker for determining whether the data can be subjected to quantitative analysis; two executor for quantitative analysis, and two other executor for qualitative analysis.

Configuring Node Tasks: For the data collection executor, we configure it to obtain data from a specific input file. For the decision maker, we configure it to determine the collected data through prompting LLM and set a rule: if LLM returns an affirmative answer, trigger the "YES" path; otherwise, trigger the "NO" path.

The DSL file for this workflow is provided in the appendix section, demonstrating how the workflow is defined and configured using JSON format. From it, you would see that in the *input_parameters* element of each node, there is an item in the type of *prompt_template*, which is corresponding to a text file. The content of this text file is the template for the prompt of the current node to prompt LLMs.

The text content is as follows:

*I am a ${role} at a ${organization}, and I am now providing a ${report} for my ${customer} client. The information of ${target} for ${goal} is between the two "---" below, and the data part between the two "***", please judge whether it is possible to conduct a quantitative analysis of ${target} based on the original information and the data therein. If yes, please input "yes", otherwise, please output "no":*

*---*

*${content}*

*---*

*\*\*\**

*${data_reader_output}*

*\*\*\**

Each *${XXX}* is a specific template parameter, and the values of these parameters are divided into two parts. Some common parameters are stored separately in a JSON file, such as "*${GF_ROOT}/data/workflows/Ads/input/parameters/trend.json*" in the example above. Its content is as follows:

```
{
    "organization": "Large Advertising Company",
    "role": "Marketing Director, whose responsibility is to play a core role in the writing of marketing plans, propose strategies and insights based on customer needs, guide creative and delivery execution, and be responsible for the final quality of the plan",
    "customer": "Coffee Producer",
    "report": "Marketing Plan",
    "goal": "Packaged Coffee",
    "target": "Industry Trends",
    "action": "Extract insights and describe the thought process of insights",
    "rules": "Note: Only need to provide the thought process, no need to give a plan. The thought process should start with a serial number and not be summarized at the end",
    "suggest": "",
    "content": "New coffee brands and products are emerging in China, and the market is severely saturated. However, there has been no innovation in the coffee market in Europe and America for decades, which is a new opportunity for packaged coffee. With the decrease in production in some coffee-producing countries and the further increase in coffee demand in various countries, the global coffee market supply gap pushes coffee prices to new highs, which also becomes an opportunity for Chinese packaged coffee brands to go overseas. Emerging instant coffee brands in China, such as Coffee Brand-1, Coffee Brand-2, and Coffee Brand-3, have seized this opportunity and entered the US e-commerce platform,
```

```
taking the lead in grabbing market share in the
United States."
}
```

If there are other specific parameter values for each node, they are defined in the DSL file's *input_parameters* element.

## VI. CASE STUDY: MARKETING STRATEGY PLAN GENERATION WORKFLOW

To demonstrate how GoTFlow's capability in handling complex business requirements, we will explore this through an example of automating the generation of marketing strategy reports.

In this case study, GoTFlow was utilized to automate the process of generating marketing strategy reports. Typically, this process involves multiple steps including the collection of market data, analysis of consumer behavior, assessment of the effectiveness of marketing channels, and the creation of a comprehensive strategy report based on this information.

When designing this workflow, the entire process was divided into two stages:

### A. First Stage: Data collection, analysis, and insight generation.

In this phase, data corresponding to four distinct parts (as shown in the diagram below) is collected, analyzed, classified, and clear inference objectives are determined. Then, based on the inference objectives for each category of information, insights are generated.

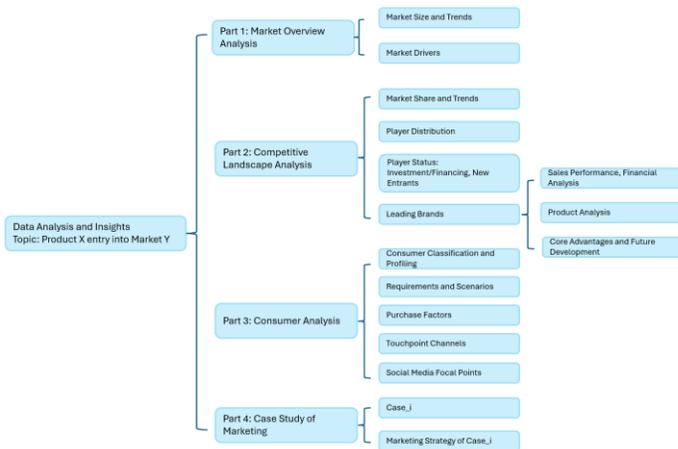

Fig. 4. Tasks of the first stage.

Each part is further divided into two workflows: the former is responsible for data collection, inference objective analysis of the data, and tagging the data with various predefined type labels; the latter generates insights based on the output from the former in relation to the inference objectives.

Several executor are defined within each workflow, such as data collection, data analysis, etc., each with a clear task and output.

### B. Second Stage: Generating and polishing the market planning report.

This stage includes three steps: generation of the report theme and outlines for each section; creation of strategic recommendations; and report refinement. Each step corresponds to a workflow.

Of course, in reality, all the above workflows in both stages could be integrated into a single workflow, but for ease of debugging and flexibility of generation, they were split into more than twenty relatively short workflows, which also reflects the flexibility of GoTFlow.

GoTFlow successfully completed the generation of the entire marketing strategy report. The automation of data collection and processing significantly shortened the required time for the whole process and also improved the accuracy of data handling. The generated report included not only a comprehensive market analysis but also provided data-driven specific marketing recommendations.

In the case study, the integration of LLMs and GoTFlow allowed for dynamic adjustments of workflows based on real-time data and situational changes. For instance, the system could modify the marketing strategy recommendations based on the latest market trends and consumer behavior insights, ensuring that the generated report remains relevant and up-to-date.

In this way, GoTFlow is proven to be able to improve work efficiency and provide solid support for data-driven decision-making as well.

## VII. DISCUSSION

By integrating LLMs, GoTFlow permits the dynamic adjustment of workflows, enabling modifications in response to real-time data and situational changes, and provides more intelligent and efficient data processing and decision support than most conventional tools offer.

Beyond its use in the generation of marketing strategy reports, GoTFlow is also applicable to various other scenarios.

For example, GoTFlow can facilitate the automation of task allocation and progress tracking for project management; it can process a vast array of customer inquiries and providing tailored responses in customer service; it can analyze intricate logistical data, thereby optimizing inventory management and distribution planning, for supply chain management…

Despite its numerous advantages, GoTFlow still faces certain challenges:

1. Complex Workflow Management: Managing intricate workflows with multiple decision points and branching paths presents a challenge in ensuring efficient and error-free execution.

2. Scalability: As workflows grow in complexity and size, ensuring that GoTFlow can scale to handle increased demands without sacrificing performance is crucial.

3. Dynamic Adaptation: Keeping up with changing requirements and conditions within a workflow

requires GoTFlow to continuously adapt, which can be challenging in rapidly evolving business environments.

4. User-Friendliness: Ensuring that GoTFlow remains accessible and easy to use for users with varying levels of technical expertise is a challenge, especially as the system's capabilities expand.

To address these challenges, a few future development directions would be possible:

1. Enhanced AI Integration: Incorporating more advanced AI and machine learning techniques to improve decision-making nodes and automate more complex tasks within workflows.

2. User Interface Improvements: Developing a more intuitive user interface or graphical workflow designer could help users more easily construct and manage their workflows.

3. Customization and Extensibility: Allowing users to create and integrate custom node types or extensions could enhance GoTFlow's flexibility and applicability to a wider range of tasks.

4. Performance Optimization: Continuous efforts to optimize the underlying architecture for better performance and scalability, ensuring GoTFlow can handle larger and more complex workflows efficiently.

5. Collaboration Features: Adding features that facilitate collaboration among multiple users or teams could enable more complex workflows that require input and decision-making from different stakeholders.

6. Cross-Platform Compatibility: Ensuring GoTFlow works seamlessly across different platforms and environments, making it more accessible to users regardless of their preferred technology stack.

It is believed that with ongoing development and refinement, GoTFlow can continue to evolve as a leading solution for automating and optimizing complex workflows, further solidifying its position as an innovative tool in the field of workflow automation and AI.

APPENDIX: THE DSL FILE FOR MARKETING PLANNING WORKFLOW

```json
{
  "output_dir_path": "${GF_ROOT}/data/workflows/Ads/output",
  "input_parameters":[
    {
        "suffix": "trend",
        "file_path": "${GF_ROOT}/data/workflows/Ads/input/parameters/trend.json"
    }
  ],
  "flow_items": [
  {
    "id": "data_reader",
    "description": "read data from the result of task_a",
    "type": "executor",
    "input_parameters": [
      {
        "name": "prompt_template_file_path",
        "type": "prompt_template",
        "file_path": "${GF_ROOT}/data/workflows/Ads/prompts/sum_data_reader.txt"
      }
    ],
    "output": [{
      "type": "variable",
      "name": "data_reader_output"
    }],
    "next_nodes": ["determine_data_feature"]
  },
  {
    "id": "determine_data_feature",
    "description": "",
    "type": "decision_maker",
    "input_parameters": [
      {
        "name": "prompt_template_file_path",
        "type": "prompt_template",
        "file_path": "${GF_ROOT}/data/workflows/Ads/prompts/sum_data_feature_determine.txt"
      },
      {
        "name": "data_reader_output",
        "type": "output_variable"
      }
    ],
    "output": [{
      "type": "variable",
      "name":  "is_quantitative_data"
    }],
    "condition": {
      "is_composed": false,
      "data_source": {
        "type": "output_variable",
        "name": "is_quantitative_data"
      },
      "operator": "equal",
      "operand": "yes"
    },
    "forward_paths": [
      {
        "condition_result": true,
        "next_nodes": ["data_trend_miner"]
      },
      {
        "condition_result": false,
        "next_nodes": ["qualitative_analysis_1", "qualitative_analysis_2"]
      }
    ]
  },
  {
    "id": "data_trend_miner",
    "description": "",
    "type": "executor",
    "input_parameters": [
      {
        "name": "prompt_template_file_path",
        "type": "prompt_template",
        "file_path": "${GF_ROOT}/data/workflows/Ads/prompts/sum_trend_miner.txt"
      },
      {
```

```json
      "name": "data_reader_output",
      "type": "output_variable"
    }
  ],
  "output": [{
    "type": "variable",
    "name":  "data_trend_miner_output"
  }],
  "next_nodes": ["quantitative_analysis"]
},
{
  "id": "quantitative_analysis",
  "description": "",
  "type": "executor",
  "input_parameters": [
    {
      "name": "prompt_template_file_path",
      "type": "prompt_template",
      "file_path": "${GF_ROOT}/data/workflows/Ads/prompts/sum_quantity_analysis.txt"
    },
    {
      "name": "data_trend_miner_output",
      "type": "output_variable"
    }
  ],
  "output": [{
    "type": "file",
    "name":  "quantitative_analysis_output.txt"
  }],
  "next_nodes": []
},
{
  "id": "qualitative_analysis_1",
  "description": "",
  "type": "executor",
  "input_parameters": [
    {
      "name": "prompt_template_file_path",
      "type": "prompt_template",
      "file_path": "${GF_ROOT}/data/workflows/Ads/prompts/sum_quality_analysis_1.txt"
    },
    {
      "name": "data_reader_output",
      "type": "output_variable"
    }
  ],
  "output": [{
    "type": "file",
    "name":  "qualitative_analysis_1_output.txt"
  }],
  "next_nodes": []
},
{
  "id": "qualitative_analysis_2",
  "description": "",
  "type": "executor",
  "input_parameters": [
    {
      "name": "prompt_template_file_path",
      "type": "prompt_template",
      "file_path": "${GF_ROOT}/data/workflows/Ads/prompts/sum_quality_analysis_2.txt"
    },
    {
      "name": "data_reader_output",
      "type": "output_variable"
    }
  ],
  "output": [{
    "type": "file",
    "name":  "qualitative_analysis_2_output.txt"
  }],
  "next_nodes": []
}
]
}
```